\documentclass[letterpaper, 10 pt, conference]{ieeeconf}  

\IEEEoverridecommandlockouts                              

\usepackage{graphicx}
\usepackage{mathrsfs}
\usepackage{amsmath}
\usepackage{amssymb}
\usepackage{algorithm}
\usepackage[noend]{algpseudocode}
\usepackage{color}
\usepackage{subfigure}
\usepackage{bbm}
\usepackage{url}
\usepackage{hyperref}

\usepackage{caption}

\usepackage{multirow}
\usepackage{siunitx}
\usepackage{booktabs}

\title{\LARGE \bf
Active localization of multiple targets using noisy relative measurements*
}

\author{Selim Engin$^{1}$ and Volkan Isler$^{1}$
\thanks{*This work was supported by the NSF grant \#1617718.}
\thanks{$^{1}$Selim Engin and Volkan Isler are with Department of Computer Science and Engineering,
        University of Minnesota, MN 55455, USA
        {\tt\small \{engin003, isler\}@umn.edu}}%
}

\begin{document}

\maketitle
\thispagestyle{empty}
\pagestyle{empty}

\begin{abstract}
Consider a mobile robot tasked with localizing targets at unknown locations by obtaining relative measurements. 
The observations can be bearing or range measurements.
How should the robot move so as to localize the targets and minimize the uncertainty in their locations as quickly as possible?
Most existing approaches are either greedy in nature or rely on accurate initial estimates.

We formulate this path planning problem as an unsupervised learning problem where the measurements are aggregated using a Bayesian histogram filter.
The robot learns to minimize the total uncertainty of each target in the shortest amount of time using the current measurement and an aggregate representation of the current belief state.
We analyze our method in a series of experiments where we show that our method outperforms a standard greedy approach. In addition, its performance is also comparable to an offline algorithm which has access to the true location of the targets.
\end{abstract}

\section{Introduction}

\begin{figure}[ht!]
    \centering
    \includegraphics[width=.8\columnwidth]{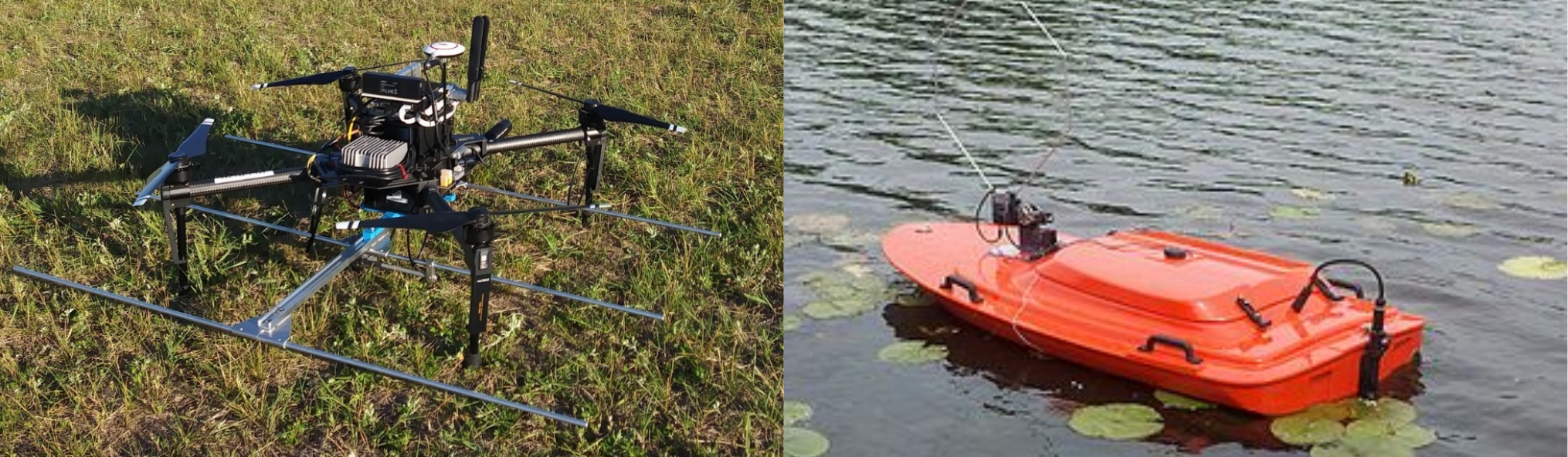}
    \caption{\textbf{Target localization problem:} given noisy relative measurements, the goal is to localize the targets as quickly as possible using a robot with complex dynamics such as an autonomous quadrotor (left) or surface vehicle~\cite{vander2015algorithms} (right).}
    \label{fig:concept}
\end{figure}

Environmental monitoring is an application area where robotics can have a major impact.
Robots can be used for gathering data, collecting samples and performing surveillance across large environments over long periods of time.
A practical problem of interest is the localization of animals in the wild who have been previously radio-tagged by wildlife biologists.
The animals can be localized by mobile robots which can obtain bearing measurements using a directional antenna. Figure~\ref{fig:concept} shows two examples of such platforms built by our group.

In this paper, we study the problem of localizing multiple targets given their noisy measurements relative to a robot. In particular, we study the bearing-only and range-only measurement models which are commonly used in practice.
For both of these models, the localization uncertainty given the true target location and robot measurement locations is well established (Section~\ref{sec:uncertainty}). However, computing the measurement locations is hard when the true location of the target is known or if there are multiple targets.

Existing methods described in Section~\ref{sec:relwork} mostly focused on the case where there is only a single target. 
Moreover, most of them are greedy in nature, optimizing the mutual information or entropy at each step locally. This approach is however susceptible to getting stuck in local minima when the best next steps are in opposite directions.
Other approaches rely on accurate initial estimates, which may not be available in general settings.
We show that instead of using a local greedy approach, long-term planning and maximizing the expected rewards can help the localization system.
We assume all the targets to be initially observable, therefore we do not address the search problem. However, a noisy initial observation of the target can result in arbitrarily bad initial estimates.
Our method is able to accurately localize targets using a bearing-only sensor with noise level up to $30^{\circ}$ angles, and range measurements with up to $30$ meters of error in a $200 \times 200$ $m^2$ area.

Our work lies in the class of model-free Reinforcement Learning (RL), where the robot learns to accomplish a given task in a fixed time-horizon without using prior information about the environment or robot dynamics.
In contrast to model-based methods, this approach does not use an engineered or a learned dynamics model. Instead, it directly optimizes its policy through a sequence of trial-and-error process~\cite{sutton2018reinforcement}.

While the policy learning is performed in a model-free paradigm, our method uses a representation of the environment that encapsulates the sensor modality and the uncertainty of the estimations.
Specifically, we use a top-down image of the environment where the intensity of each pixel indicates the likelihood of that point being the position of the target.
As the sensor noise increases there are more pixels indicating a high likelihood of containing the target.


\section{Related Work and Contributions}
\label{sec:relwork}
The problem of localizing targets has received significant attention and been studied in several different settings. In this section we go over three different scenarios
where each setup has a different set of assumptions about the target and sensor mobility: 
a) sensor placement, the problem of deploying a group of static sensors to localize static targets; b) active target localization, the problem of localizing static targets with mobile sensors; and c) active target tracking, the problem of tracking mobile targets with mobile sensors. 


\subsection{Sensor Placement}
In the sensor placement problem we have a group of static targets whose positions are unknown residing in a given candidate area and an uncertainty model.
The goal is usually to minimize the number of sensors deployed in the area given a desired uncertainty level~\cite{guestrin2005near, tekdas2010sensor, chakrabarty2002grid}, or to optimize the sensor-target geometries after fixing the number of sensors~\cite{zhao2012optimal}.
Both the targets and sensors are assumed to be static in this setting, and a greedy approach is shown to have a near-optimal performance~\cite{guestrin2005near}.
The results in this setting, however, are not directly applicable to the case when the targets or sensors are moving. 

\subsection{Active Target Localization}
In the problem of active target localization, one or multiple mobile robots need to localize the targets as quickly as possible. In~\cite{ponda2009trajectory} a method is proposed to perform target localization with bearing measurements using the Fisher information to optimize the trajectory of a single aerial vehicle. In~\cite{vander2015algorithms} a cooperative algorithm is presented to localize a static target using two networked robots with communication constraints. The robots in this work need to periodically meet to update their estimate of the target position.

\subsection{Active Target Tracking}
In the active target tracking problem, both the sensors and targets are able to move.
The target motion can be modeled in various ways and this model plays an important role in the performance of the tracking system. 
One approach is to treat the target motion adversarial, which means that the target actively evades the sensor~\cite{battistini2014differential, shima2002time}.
Another approach can model the motion of the target to be a random walk, especially when there is no prior information about the mobility of the target~\cite{schulz2001tracking, yu2005decentralized}.
The mobility is modeled by a 2D Brownian motion in~\cite{nishimura2018sacbp} where the target can move within a bounded region.

For this problem setting, most of the existing approaches use a position-based measurement model~\cite{jeong2019learning, he2010efficient, hausman2015cooperative, pierson2017distributed}. This enables the robot to estimate the target positions with a few observations, unlike using bearing or range measurements where the geometry of the sensor-target configuration significantly impacts the accuracy of the estimator.




\subsection{Statement of contributions}
Our contributions can be summarized as follows.
\begin{itemize}
    \item We propose a novel RL formulation of the multi-target localization problem.
    \item We present a representation to model the localization estimates for each target encapsulating the uncertainty in the predictions.
    \item We compare our RL algorithm against a greedy approach that uses local observations to plan the sensor motion, and an offline algorithm which has access to the true positions of the targets.
\end{itemize}

\section{Preliminaries}
The setup in our problem consists of a tracker robot with an omnidirectional bearing or range sensor and a set of $m$ targets deployed in an environment $\mathcal{V}$. The mobility of the robot is modeled to be holonomic, that is it can do instantaneous turns without any kinematic constraints.
We denote the position of the robot at time $t$ by $p_t \in \mathbb{R}^2$.
Similarly, the locations of the targets are $\mathbf{q} = \{q_1, \dots, q_m \}$, where $q_i \in \mathbb{R}^2$, and the predicted positions of the targets are $\hat{\mathbf{q}}$. 
The trajectory of the robot until time step $t$ is denoted by $P_{1:t} = p_1, \dots, p_t$. 
Throughout the paper we use a global frame, such as the initial coordinate frame of the robot to denote the positions of the robot and targets. 

The true bearing direction from a sensor location $p = (p^x, p^y)$ to a target location $q_{i}  = (q_i^x, q_i^y)$ is denoted by $\phi_{i} = \phi(p,q_i)$ and expressed as:
\begin{equation}
    \phi_{i} = \arctan \big((q_{i}^y - p^y) / (q_{i}^x - p^x) \big).
\end{equation}

We assume that the measurement model has a Gaussian distribution noise.
Then, a bearing measurement from a sensor can be written as $\hat{\phi}_i = \phi_i + e$, where $e \sim \mathcal{N}(0, \sigma_s^2)$ and $\sigma_s^2$ is the variance of the sensor measurement error.
We denote the set of bearing measurements acquired from a robot position $p_t$ at time step $t$ by $\hat{\Phi}_t = \{\hat{\phi}_1, \dots \hat{\phi}_m\}$.

The distance between a sensor $p$ and a target $q_i$ is denoted by $d_i = d(p, q_i)$. Similar to the bearing measurement model, we assume a normally distributed zero-mean noise for the range measurements. A range measurement at time $t$ is expressed as $\hat{d}_i = d_i + e$, where $e \sim \mathcal{N}(0, \sigma_s^2)$.
Throughout the paper we denote the relative noisy measurements by $\hat{z}_i$, and use $\hat{\phi}_i$ or $\hat{d}_i$ to indicate a specific sensor model.

\subsection{Uncertainty Measures}
\label{sec:uncertainty}
In this section we describe the uncertainty measures used in the paper.

\subsubsection{Fisher Information}
Predicting a target location using relative measurements is typically achieved by triangulation from multiple positions.
Quantifying the accuracy of the triangulation relies on the geometry of the sensor-target configuration and the noise in the sensor measurement.

For a given sensor-target geometry and a noise level in the measurements, we can use the Fisher Information Matrix (FIM) to compute the amount of information given by the measurements.
The FIM characterizes the amount of information an observable parameter carries about an unobservable variable.
In our case, the observable parameter is the relative measurement and the unobservable variable of interest is the position of the target.

Suppose $\mathcal{F}_q$ is the FIM for a given sensor-target configuration, where $q^*$ is the true location of the target. The ($i,j$)-th entry of $\mathcal{F}_q$ for a single target can be expressed as: 

\begin{equation}
    (\mathcal{F}_q)_{i,j} = \mathbb{E}\Big[\frac{\partial}{\partial q_i} \ln(\mathbb{P}(\hat{z}|q)) \;\; \frac{\partial}{\partial q_j} \ln(\mathbb{P}(\hat{z}|q))\Big],
\end{equation}

where $\mathbb{P}(\cdot)$ denotes the probability density function.
The determinant of the FIM is inversely proportional to the uncertainty area of the estimation~\cite{van2004detection, bishop2007optimality}.
The determinant of $\mathcal{F}_q$ for $n$ bearing measurements is given by~\cite{bishop2010optimality}:
\begin{equation}
    det(\mathcal{F}_q) = \frac{1}{\sigma_s^4} \sum_{\{p_i, p_j\} \in \mathcal{P}} {\frac{\sin ^2 (\phi(p_i, q^*) - \phi(p_j, q^*)) }{d(p_i, q^*)^2 \cdot d(p_j,q^*)^2}}, \;\; i < j
\end{equation}

where $\sigma_s^2$ denotes the variance of the measurement error, and $\mathcal{P}$ defines the set of all pairwise combinations as $\{\{p_i, p_j\}\}$ with $i, j \in \{1, \dots, n\}$ and $i < j$. 
The inverse FIM $\mathcal{F}_q^{-1}$ (also known as the Cramer-Rao inequality lower bound) quantifies an uncertainty ellipsoid for unbiased estimators.
The square root of the eigenvalues of $\mathcal{F}_q^{-1}$, denoted by $\sqrt{\lambda_i}$ defines the axis length of the ellipsoid along the $i$-th dimension.

The determinant of $\mathcal{F}_q^{-1}$ provides a scalar for measuring the uncertainty.
For a given sensor-target geometry and a measurement model, we measure the total uncertainty by:

\begin{equation}
    \mathcal{U}(P, \mathbf{q}) = \sum_{i=1}^m {det(\mathcal{F}_{q_i}^{-1})}.
\end{equation}

Since the true position of the target is required to compute the Fisher information, it is challenging to obtain an accurate uncertainty estimate based on the Fisher information in practical scenarios. 
The next uncertainty measure we present is better suited for practical applications where we do not have access to the true positions. We use the Fisher information as an uncertainty estimate in the offline case where the true position of the targets are available to the algorithm.


\subsubsection{Bayesian Histograms}

In Bayesian histograms we discretize the environment $\mathcal{V}$ into a grid, where each cell of the grid indicates the likelihood of being the target position given the measurements.
Bayesian histograms are particularly useful to represent the likelihood posteriors for models with nonlinear dynamics.
Filtering techniques relying on linearization, in general are sensitive to disturbances when using a nonlinear sensor model like bearing measurements.
Without accurate initialization, the measurement errors can easily lead to a poor localization performance.
Several existing methods~\cite{hoffmann2019rollout, cliff2015online} have also used Bayesian histograms for the problem of target localization. 

Since the measurement model assumes a zero-mean normal distribution for the sensor noise, we can compute the probability of a point $v$ being the true target location $q^*$ given a relative measurement $\hat{z}$ as:

\begin{equation}
\mathbb{P}(v = q^* | p, \hat{z}) = f(z_v | \hat{z}, \sigma_s^2) = \frac{1}{\sqrt{2\pi\sigma_s^2}} \exp\{{-\frac{1}{2\sigma_s^2}(z_v-\hat{z})^2}\},
\end{equation}

where $z_v$ is the true angle $\phi(p, v)$ or distance $d(p, v)$ between the point $v$ and a robot position $p$. 
Over the course of the robot's trajectory, the likelihood of a point is updated as measurements are obtained:
\begin{equation}
\mathbb{P}(v = q^* | P_{1:T}, \hat{z}_{1:T}) = \prod_{k=1}^T f(z(p_k, v) | \hat{z}_k, \sigma_s^2).
\end{equation}

\begin{figure}[ht!]
    \centering
    \includegraphics[width=.9\columnwidth]{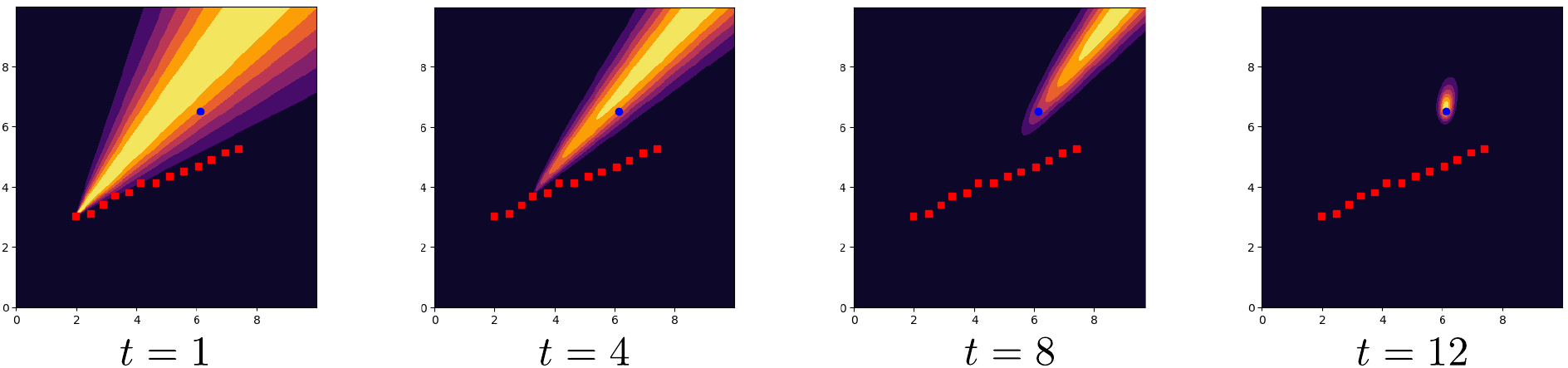} 
    \caption{The Bayesian histograms computed by aggregating the bearing measurements taken from the sensor trajectory shown in red. The blue dot corresponds to the true target location, and the darker to brighter colors in the heatmap indicate an increase in the likelihood.}
    \label{fig:heatmap}
\end{figure}

At each measurement update, the probabilities are normalized such that $\max_{v \in \mathcal{V}} f(z(p_k, v) | \hat{z}_k, \sigma_s^2) = 1$, where $\mathcal{V}$ is a given domain for the target positions. 
When initializing the heatmap, we use a uniform distribution of probability 1 indicating that each point in the domain is equally likely to be the true position of the target.
We denote the histogram at time step $t$ for a target $q_i$ as $\mathcal{H}_t(P_{1:t}, \hat{\mathbf{z}}_{1:t}, i)$. The histograms are updated separately and stacked as a tensor $\mathcal{H}_t(P_{1:t}, \hat{\mathbf{z}}_{1:t}) \in \mathbb{R}^{m \times W \times H}$.
The histogram dimensions $W \times H$ determine the granularity of the grid discretization of the environment and result in a trade-off between accuracy and efficiency. Finer resolution histograms have smaller discretization errors at the cost of higher computational resources.
An instance of the Bayesian histogram representation is shown in Figure~\ref{fig:heatmap}. At each time step the bearing measurement is incorporated to update the likelihood of the points in $\mathcal{V}$.
The sensor trajectory and target position are overlaid in the heatmap figures to give context.

\subsection{Reinforcement Learning}
In Reinforcement Learning (RL), we have an agent sensing and acting within an environment over a course of discrete time steps. 
At each time step, the agent observes the current state of the environment and performs an action. Depending on the current state and action pair, the state changes and the agent receives a reward according to an environment model~\cite{sutton2018reinforcement}. 

These type of problems are usually formulated as Markov Decision Processes (MDP) described by states $s \in \mathcal{S}$, actions $a \in \mathcal{A}$, transition dynamics $\mathcal{T}: \mathcal{S} \times \mathcal{A} \times \mathcal{S} \rightarrow [0,1]$, a reward function $r: \mathcal{S} \times \mathcal{A} \times \mathcal{S} \rightarrow \mathbb{R}$, and a discount factor $\gamma \in [0,1]$.
For the case when the agent cannot fully observe the states, this description is redefined as a Partially Observable MDP (POMDP), where there is also an associated observation function.

The agent selects actions using a policy $\pi(a|s)$ so as to maximize the expected $\gamma$-discounted future rewards given by $G_t = \mathbb{E}\big[\sum_{i=t}^T \gamma^{i-t} r(s_i, a_i, s_{i+1})\big]$, where $T$ is the planning time-horizon.
Suppose that $Q^\pi (s,a)$ is the expected future returns by taking the action $a$ in state $s$ using the policy $\pi$, which is expressed as $Q^\pi (s,a) = \mathbb{E}\big[G_t|s_t=s, a_t=a \big]$.
The goal of RL algorithms is to learn an optimal policy $\pi^*$ that maximizes the expected future returns for all state and action pairs.
A popular method for this problem is $Q$-learning~\cite{watkins1992q}, where the Bellman equation,

\begin{equation}
\label{eq:bellman}
    Q(s_t,a_t) = \mathbb{E}_{\mathcal{T}(s_{t+1}|s_t, a_t)} \big[r(s_t,a_t,s_{t+1}) + \gamma\max_{a'} Q(s_{t+1}, a') \big],
\end{equation}
is recursively used to adjust $Q^\pi$ to approach the optimal $Q$-values for $\pi^*$.
When the state and action spaces are large, this value function can be approximated as a neural network $Q_\theta(s,a)$, with parameters $\theta$.


\section{Problem Statement}
We are now ready to formally state the problem studied in this paper.

\textbf{Problem 1} \textit{(Target Localization)}.
We are given a mobile robot equipped with a sensor obtaining relative measurements according to a known measurement model, a starting position $p_1$, a fixed planning horizon $T$ and a set of targets whose positions are unknown. 
In addition, we are given a function $\mathcal{U}(P, q)$ measuring the uncertainty of the target estimates for a given sensor-target geometry and a measurement model.
The goal is to find a path $P_{1:T} = p_1, \dots, p_T$ for the robot such that the total uncertainty of the target positions $\sum_{i=1}^m \mathcal{U}(P_{1:T}, q_i)$ is minimized.


\section{Method}
Our method uses two different models to represent the state.
In this section we go over these representations and present our unsupervised learning method for target localization.

\begin{figure}[ht!]
    \centering
    \includegraphics[width=0.95\columnwidth]{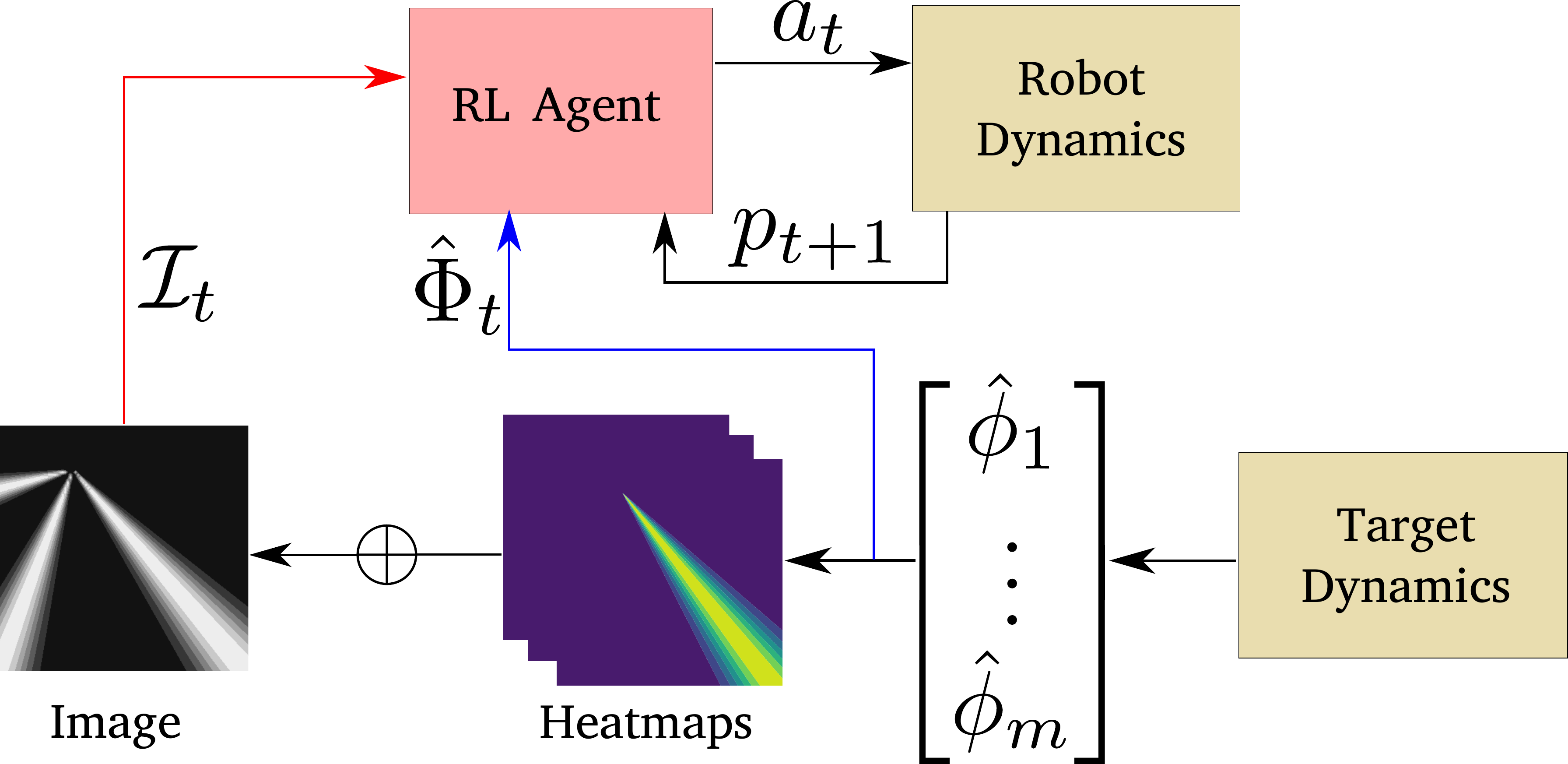}
    \caption{\textbf{Method overview:} A single iteration of the robot-environment interaction. The blue arrow indicates the relative measurements vector used in the multi-modal representation, while the red arrow marks the image used as the observation in the image representation.}
    \label{fig:overview}
\end{figure}

An overview of our method is shown in Figure~\ref{fig:overview}.
At every iteration, the robot acquires noisy measurements from the target locations according to the measurement model.
The measurements are either taken as raw observations, or they are aggregated into a single image representation. 
Depending on the observation, the agent selects an action which drives the robot to a new position. The measurements are accumulated in the form of heatmaps at each step, which are then used to predict the position of the target.
This process repeats until the end of the planning time-horizon.

\subsection{Representing the observations}
We start with describing the representations we use for the target observations.

\subsubsection{Multi-modal representation}
The first representation our method uses is a multi-modal Gaussian distribution for characterizing the localization belief.
At each step, the robot receives a distinct bearing or range measurement from all the targets (i.e. perfect assignment).
The sensor noise $\sigma_s$ is assumed to be known, therefore we are able to represent the likelihood distribution over the domain $\mathcal{V}$.
We construct the likelihood of each point $v \in \mathcal{V}$ as the normalized probability $f(z_v | \hat{z}, \sigma_s^2)$, where $\hat{z}$ is the observation for a single target.


In a standard Markov process, at any time step the next state depends on the current state and the action. For localizing the targets, the robot needs to maintain an estimate of the target position using observations from the past locations.
Hence, we append the predictions for the target positions to the state vector so as to select the next actions.

The augmented state of the system is defined as $s_t := (p_t, \hat{\mathbf{z}}_t, \hat{\mathbf{q}}_t) \in \mathbb{R}^{2+m+2m}$, concatenating the position vector of the robot $p_t$, observations from each target $\hat{\mathbf{z}}_t$, and a flattened vector of the predicted positions of the targets $\hat{\mathbf{q}}_t$.
In the multi-modal representation, we use the noisy bearing $\hat{\phi}_t$ or range measurements $\hat{d}_t$ as the observation vector $\hat{\mathbf{z}}_t$.

\subsubsection{Image representation}
We also use images to represent the belief for the predicted position of the target.
The image $\mathcal{I}_t \in \mathbb{R}^{W \times H}$ at time step $t$ is computed by the sum of histograms maintained for each target estimate, expressed as:

\begin{equation}
    \mathcal{I}_t = \sum_{i=1}^m {\mathcal{H}_t(P_{1:t}, \hat{\mathbf{z}}_{1:t}, i)}.
\end{equation}

The image is then normalized to have values in the range $[0, 1]$, and resized into a desired resolution with bilinear interpolation.
We use a convolutional neural network with the ResNet-34 architecture~\cite{he2016deep} followed by three fully connected layers to encode $\mathcal{I}_t$ into a latent vector $l_t \in \mathbb{R}^c$. For all our experiments, we set the dimension of the latent code to be $c = 128$. 
This encoding vector implicitly describes the uncertainty level and the position estimates for all the targets in a latent space.

The state vector we use in the image representation is expressed as $s_t := (p_t, l_t) \in \mathbb{R}^{2+c}$, a concatenation of the position vector $p_t$ and latent code $l_t$.
The RL algorithm we present next inputs the state vector corresponding to the desired observation representation.

\subsection{Learning target localization}
We build our method on the Twin Delayed Deep Deterministic (TD3) algorithm, a deep reinforcement learning algorithm introduced in~\cite{fujimoto2018addressing} that uses policy gradients to directly optimize the policy network.
Similar to the Deep Deterministic Policy Gradient algorithm~\cite{lillicrap2015continuous}, TD3 uses an actor-critic architecture that jointly learns an actor $\pi_\varphi$ (the policy) and a critic function $Q_\theta$ (the value function) modeled by neural networks with parameters $\varphi$ and $\theta$, respectively.

A replay buffer is maintained to store the previous interactions of the agent, as in the Deep Q-Network algorithm~\cite{mnih2015human}.
The replay buffer stores samples in the form of ($s_i$, $a_i$, $s_{i+1}$, $r_i$, $y_i$) tuples, where each sample is associated with a state transition from $s_i$ to $s_{i+1}$ by executing the action $a_i$. With this transition, the agent receives a reward $r_i$, and the termination of the episode is indicated by a label $y_i \in \{0,1\}$.
To prevent the over-estimation bias of the value function, TD3 trains two (twin) critic networks that are updated at different (delayed) frequencies and uses the minimum of the two as the value estimate.
In our notation, $\min_{j=1,2} Q_{\theta_j}(s,a)$ is replaced by $Q_\theta(s,a)$ for brevity.

For both the actor and critic, periodically updated target networks $\pi_{\varphi'}$ and $Q_{\theta'}$ are also used. Constraining the change in the target networks to be slow helps the stabilization when learning the actor and critic functions.

The actor and critic networks are trained with the Adam optimizer~\cite{kingma2014adam} using the loss functions $L_{actor}$ and $L_{critic}$ over mini-batches of size $B$.
The loss functions are given by:

\begin{align}
    \begin{split}
        L_{actor} ={}& \frac{1}{B} \sum_{i=1}^B - {Q_\theta(s_i, \pi_\varphi(s_i))}
    \end{split}\\
    \begin{split}
        L_{critic} ={}& \frac{1}{B} \sum_{i=1}^B \Big[ Q_\theta(s_i, a_i) - \Big(r_i + \gamma(1-y_i)\\
        &  \cdot Q_{\theta'}\big(s_{i+1}, \pi_{\varphi'}(s_{i+1})\big) \Big) \Big]^2.
    \end{split}
\end{align}

Intuitively, the actor network $\pi_\varphi$ is trained to maximize the Q-values generated from the action $\pi_\varphi(s_i)$ given a state $s_i$, by minimizing $-Q_\theta(s_i, \pi_\varphi(s_i))$.
On the other hand, the critic network is trained to minimize the Bellman error computed by a variation of the Bellman equation given in Equation~(\ref{eq:bellman}).
During the training process, the slowly changing target networks are assumed to characterize the optimal policy and the Q-function.
The policy $\pi_\varphi$ inputs the state vector $s_t$ according to the observation representation and outputs an action $a_t$.
This action is a heading angle in the continuous space represented in the inertial coordinate frame, taking values bounded as $a_t \in [0, 2\pi)$.

\subsubsection{Reward functions}

For learning the policy we use two different reward functions, one for each observation representation.
In the multi-modal representation, the reward is the negative mean squared error between the true and predicted target positions, expressed as $- 1/m \cdot \sum_i^m {||\mathbf{q}_i - \hat{\mathbf{q}}_i||^2}$.

For the image representation, on the other hand, we simply use the mean of the image intensities as the loss term. The reward function we use is given by $- \frac{1}{WH} \cdot \sum_{u,v} \mathcal{I}_t(u,v)$, where the pixel intensities of the image are all non-negative. 
Maximizing this reward function minimizes the likelihood of parts of the environment $\mathcal{V}$ not containing the targets, which in turn amounts to minimizing the total entropy of the Bayesian histograms.
The reward function does not use the true position of the targets and it is completely unsupervised. 

\section{Analysis}
In this section we evaluate our method with a set of experiments.
The experiments are designed around questions concerned about the performance of our method compared to several baseline approaches.

\subsection{Offline Trajectory Optimization}

We first investigate whether we can generate optimal trajectories for a robot to minimize the uncertainty, given the true position of the targets.

This problem has been addressed in two recent papers for bearing~\cite{he2019trajectory} and range-only~\cite{he2019optimal} measurement models, where they study the optimal one-step action for the sensor so as to maximize the information about the target estimate.
The cost function they provide for one-step maximization has the form: $J_\phi = \sin (\phi(p_{t-1}, q) - \phi(p_t, q))^2 / d(p_t, q)^2$,
which is used to select the best action to move the sensor from position $p_{t-1}$ to $p_t$ when using bearing measurements.
This expression closely resembles the geometric dilution of precision (GDOP)~\cite{kelly2003precision}, commonly used in navigation systems. 
The uncertainty estimate using the GDOP function for a given sensor-target geometry is given by:
\begin{equation}
    U_{GDOP} = \frac{d(p_i, q) \cdot d(p_j, q)}{|\sin (\phi(p_i, q) - \phi(p_j, q))|}.
\end{equation}

The offline algorithm we compare against is based on the minimization of the uncertainty measure $\mathcal{U}(P, \mathbf{q})$. This measure is equal to a constant times the GDOP function summed over all pairs of the measurement locations $P$.
We call this offline algorithm \textsc{Offline-Fisher} and present it in Algorithm~\ref{algo:offline}.

\begin{algorithm}
\caption{\textsc{Offline-Fisher}}
\begin{algorithmic}[1]
  \Require {$\mathcal{U}$: uncertainty function , $\mathbf{q}$: target positions, $T$: time horizon, $\delta_p$: step size}
  \Ensure {$P_{1:T}$: a trajectory for minimizing the uncertainty of the target estimates}
  \State $p_1 \leftarrow $ Starting position of the robot
  \For {$t \leftarrow 2$ to $T$}
    \State $a_t^* \leftarrow \arg\min_{a_t \in \mathcal{A}}{ \sum_{i=1}^m \mathcal{U}(P_{1:t-1} \cup \{p_t\}, q_i)}$
    \State $p_t^* \leftarrow \delta_p \cdot \big[\cos(a_t^*); \; \sin(a_t^*) \big]$; $P_{1:t} \leftarrow P_{1:t-1} \cup \{p_t^*\}$
  \EndFor
  \State \textbf{return} $P_{1:T}$
\end{algorithmic}
\label{algo:offline}
\end{algorithm}

\subsection{Localizing Static Targets}

In practical settings, the true location of the targets are not available to us, therefore the algorithm needs to operate under uncertainty.
The second method we compare against is a local greedy approach which chooses an action to minimize the conditional entropy given the past measurements of the targets, at each step.
This approach is commonly used in practice and it resembles e.g.~\cite{schwager2017multi}.

In this section we present experimental evaluations comparing our method against \textsc{Offline-Fisher} and the local greedy algorithms.
Specifically, we investigate the contribution of having expected future rewards in the cost function, in contrast to optimizing only the next step.

The environment used in our experiments is a $20 \times 20$ square area.
The robot step size is set to be $0.5$ units, and the planning time-horizon is 50 steps.
Variance of the sensor noise for bearing measurements is $\sigma_s^2 = 0.2^2$ which corresponds to about $11^\circ$, and for range measurements it is $1$.
Each unit in the histogram has a resolution of 10, thus the dimensions of $\mathcal{H}_t$ are $200 \times 200$.


\begin{figure}[ht!]
\centering     
\subfigure[Target localization with bearing measurements]{\includegraphics[width=0.9\columnwidth]{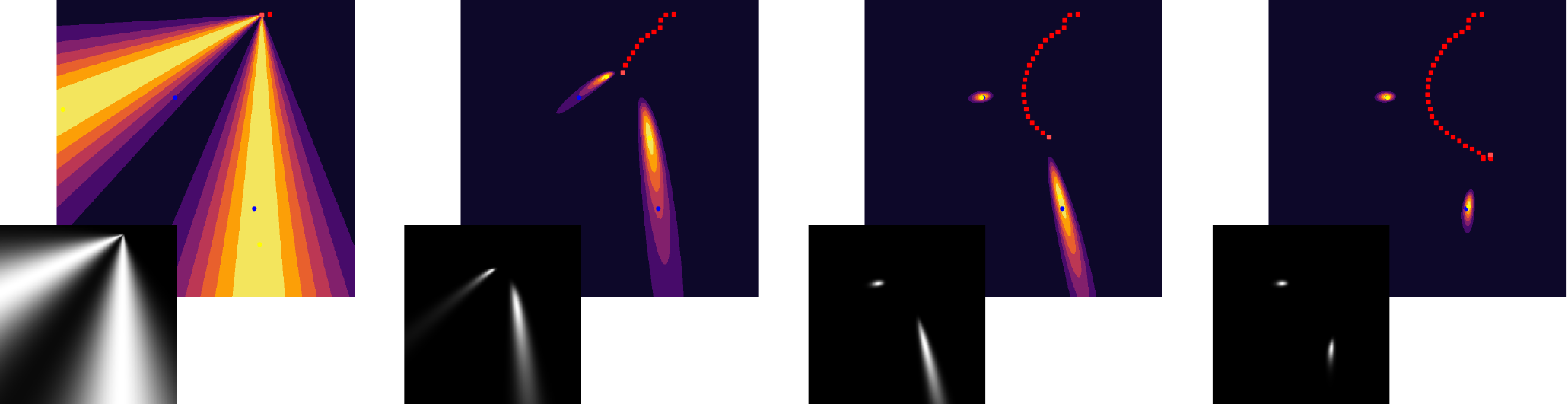}}
\subfigure[Target localization with range measurements]{\includegraphics[width=0.9\columnwidth]{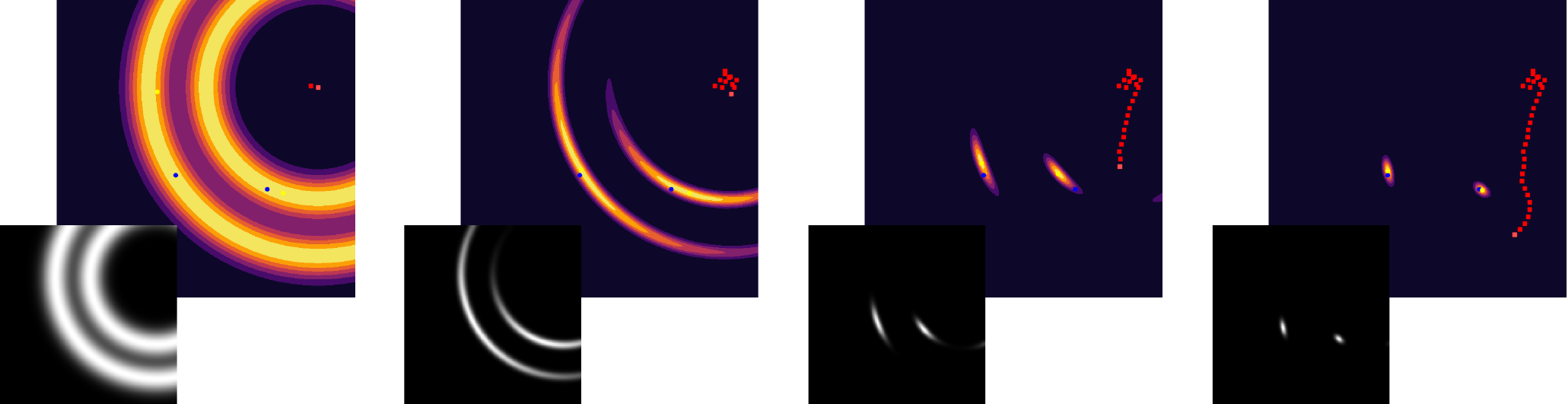}}
\caption{The trajectory of the robot acquiring bearing (a) and range (b) measurements, using the image representation. Both rows show the robot trajectory (red), true (blue) and predicted (yellow) target position, also the Bayesian histogram as the colored heatmap. The insets contain the images taken as input to the network at corresponding instants. Best viewed in color.}
        \label{fig:traj_bearing_range}
\end{figure}

Localization of two targets with bearing and range-only measurements using the image representation is shown in Figure~\ref{fig:traj_bearing_range}.
The figure shows the measurements acquired from the targets by the robot, and its trajectory to minimize the uncertainty of the predictions by using the corresponding images.
We see that the learned policy is able to generate smooth trajectories that can accurately localize multiple targets.


        
\begin{figure}[ht!]
\centering     
\subfigure[$m=4$, bearing ($\phi$)]{\includegraphics[width=0.45\columnwidth]{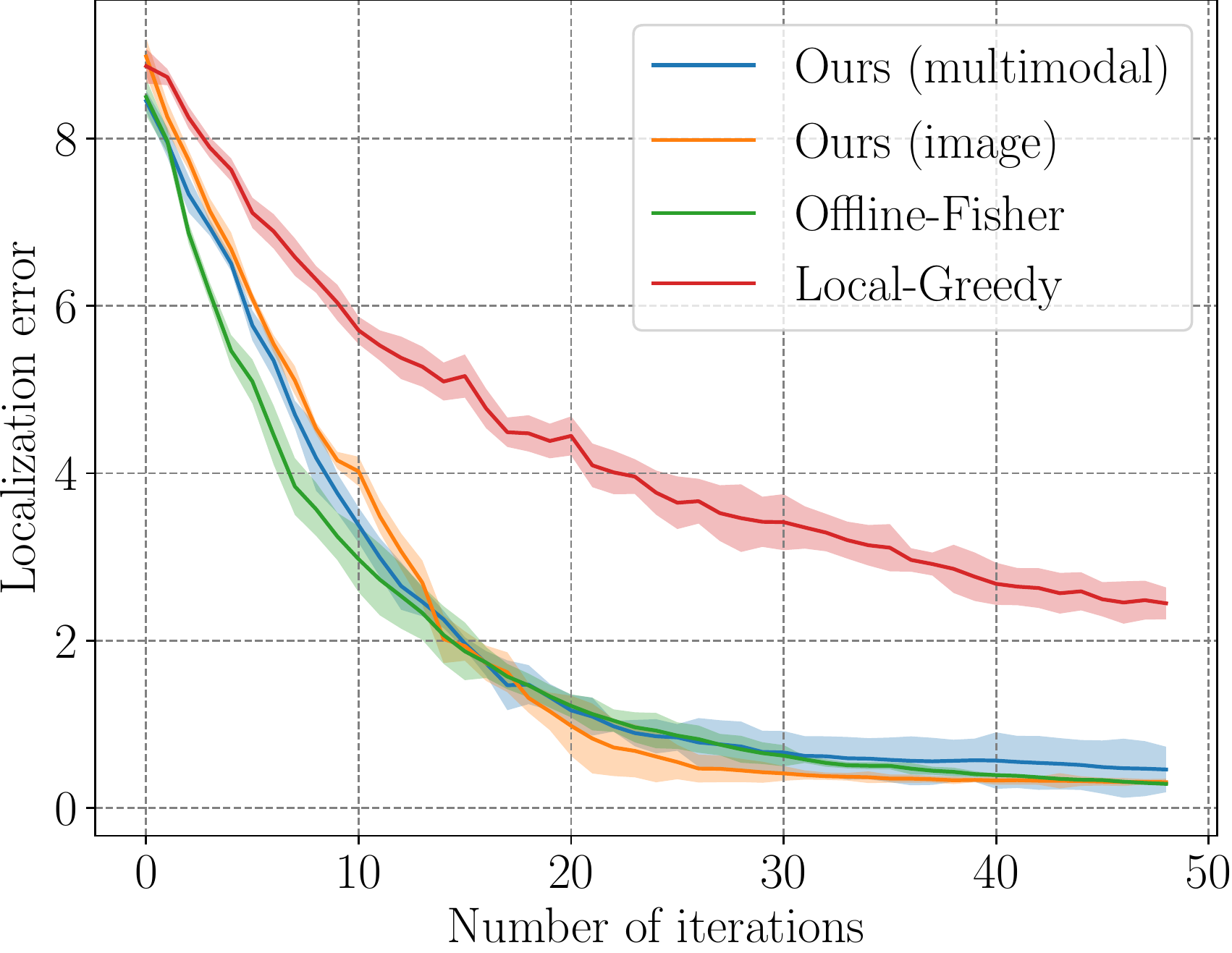}}
\subfigure[$m=8$, bearing ($\phi$)]{\includegraphics[width=0.45\columnwidth]{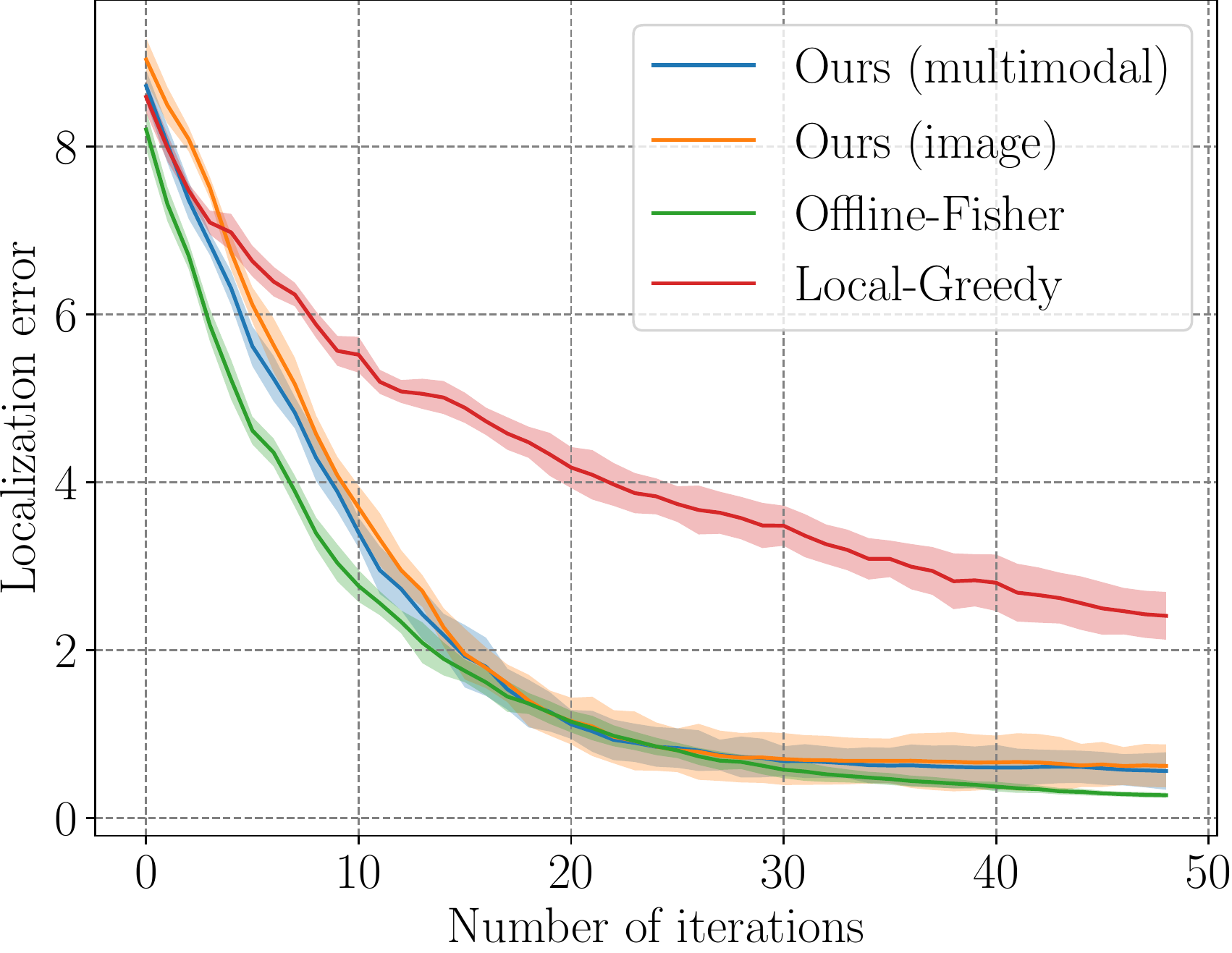}}
\subfigure[$m=4$, range ($d$)]{\includegraphics[width=0.45\columnwidth]{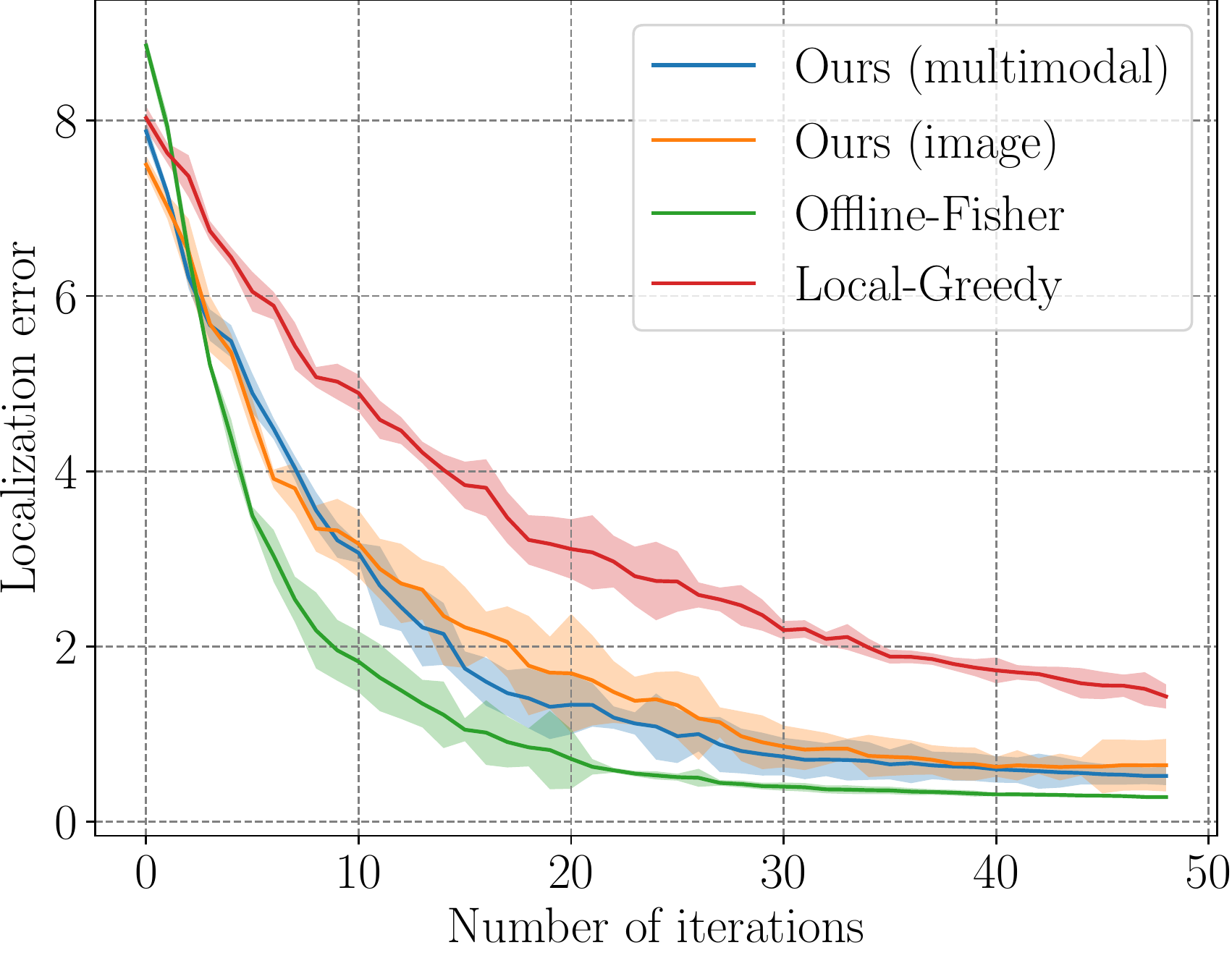}}
\subfigure[$m=8$, range ($d$)]{\includegraphics[width=0.45\columnwidth]{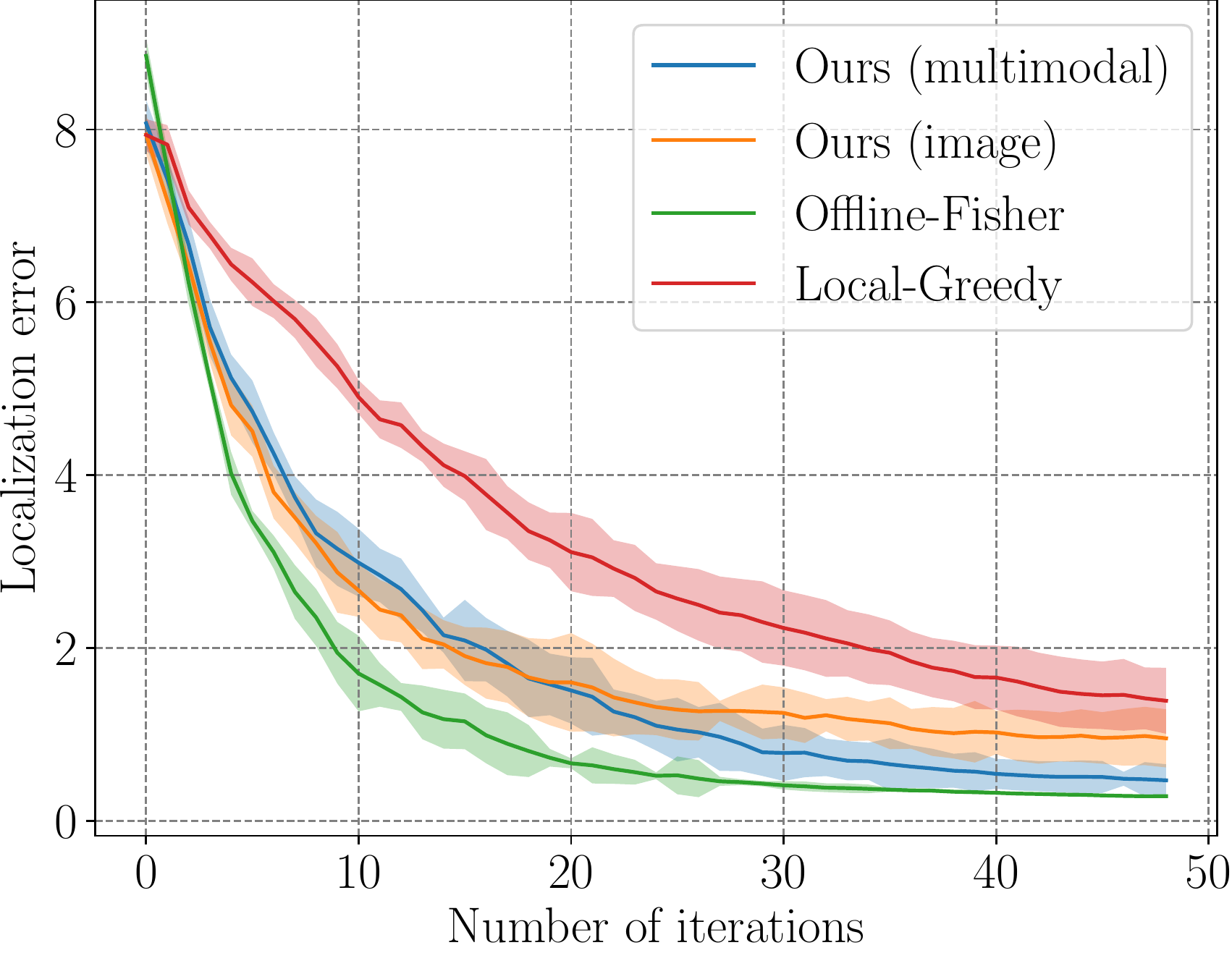}}
\caption{Mean localization errors computed over 100 trials using bearing and range measurements from $m = 4,8$ targets. Shaded area represents $1\sigma$.}
        \label{fig:locerrs}
\end{figure}

We provide quantitative results for the localizing the targets using bearing and range-only measurements in Figure~\ref{fig:locerrs}.
The plots show the mean localization error using four methods, averaged over 100 episodes.
We compute the localization error by finding the point with the highest likelihood from the uncertainty histograms, and use it to calculate the mean distance error between the predicted and true target positions.
We observe that both representations we use outperform the greedy strategy over the course of an episode, and have a performance very close to that of the offline algorithm.


The mean localization error at the end of the episodes for each method is presented in Table~\ref{table:static_error}. 
Unsurprisingly, the best trajectories minimizing the localization uncertainty are generated by the offline algorithm, in all cases.
We see that our method using either of the observation representation yields very close results to the offline algorithm.
The performance of the two representations are similar using bearing measurements, however the multi-modal representation results in better localization when using range measurements. 
A possible reason for this is the spatially characteristics of the image in the case of range measurements.
In bearing measurements the centroid of the high-likelihood pixels gives a reasonable prediction about the target position, which is not always the case for range measurements.
However, in all experiments we see that our method outperforms the greedy strategy.

\begin{table}[ht!]
\centering
\setlength\tabcolsep{6pt}
  \begin{tabular}{c l c c c}
    \toprule
 & \multicolumn{1}{l}{\textbf{Method}} & \multicolumn{1}{c}{$m=2$} & %
    \multicolumn{1}{c}{$m=4$} & \multicolumn{1}{c}{$m=8$}\\
\midrule
\multirow{4}{*}{$\phi$} & Offline-Fisher & \emph{0.228 / 0.012} & \emph{0.285 / 0.006} & \emph{0.268 / 0.301} \\
& Greedy-Local & 2.599 / 0.404 & 2.444 / 0.190 & 2.406 / 0.284 \\
& Ours (Multi-modal) & 0.481 / 0.092 & 0.457 / 0.272 & \textbf{0.557 / 0.223} \\
& Ours (Image) & \textbf{0.479 / 0.255} & \textbf{0.305 / 0.041} & 0.618 / 0.256 \\
 \midrule
\multirow{4}{*}{$d$} & Offline-Fisher & \emph{0.276 / 0.001} & \emph{0.282 / 0.007} & \emph{0.284 / 0.016} \\
& Greedy-Local & 1.357 / 0.264 & 1.430 / 0.137 & 1.388 / 0.381 \\
& Ours (Multi-modal) & \textbf{0.461 / 0.070} & \textbf{0.524 / 0.104} & \textbf{0.468 / 0.186} \\
& Ours (Image) & 0.990 / 0.257 & 0.646 / 0.299 & 0.954 / 0.336 \\
\bottomrule
\end{tabular}
\vspace{0.1cm}
\caption{Mean and standard deviation of localization errors using bearing ($\phi$) and range ($d$) measurements from static targets}
\label{table:static_error}
\end{table}

\subsection{Localizing Dynamic Targets}

We next analyze the localization performance when the targets are moving locally in a contained area.
The target movement is modeled by a 2D Brownian motion with a covariance of $0.1 I_2$, similar to the setting in~\cite{nishimura2018sacbp}.
Our current formulation of the uncertainty histogram does not allow tracking targets performing large displacements. 
However, when the target motion is relatively small our method gives reasonable performances compared to the baseline algorithms. 


\begin{table}[ht!]
\centering
\setlength\tabcolsep{6pt}
  \begin{tabular}{c l c c c}
    \toprule
 & \multicolumn{1}{l}{\textbf{Method}} & \multicolumn{1}{c}{$m=2$} & %
    \multicolumn{1}{c}{$m=4$} & \multicolumn{1}{c}{$m=8$}\\
\midrule
\multirow{4}{*}{$\phi$} & Offline-Fisher & \emph{0.696 / 0.042} & \emph{0.707 / 0.023} & \emph{0.706 / 0.039} \\
& Greedy-Local & 3.177 / 0.396 & 2.851 / 0.305 & 2.810 / 0.257 \\
& Ours (Multi-modal) & 0.981 / 0.661 & 0.992 / 0.728 & \textbf{0.899 / 0.444} \\
& Ours (Image) & \textbf{0.901 / 0.215} & \textbf{0.699 / 0.140} & 1.031 / 0.368 \\
 \midrule
\multirow{4}{*}{$d$} & Offline-Fisher & \emph{0.591 / 0.028} & \emph{0.597 / 0.013} & \emph{0.611 / 0.033} \\
& Greedy-Local & 1.985 / 0.106 & 2.003 / 0.149 & 2.159 / 0.338 \\
& Ours (Multi-modal) & \textbf{0.947 / 0.034} & \textbf{0.874 / 0.134} & \textbf{0.845 / 0.180} \\
& Ours (Image) & 1.312 / 0.328 & 0.978 / 0.279 & 1.390 / 0.195 \\
\bottomrule
\end{tabular}
\vspace{0.1cm}
\caption{Localization errors using relative measurements from moving targets}
\label{tab:dynamic_error}
\end{table}

The results of this experiment are reported in Table~\ref{tab:dynamic_error}.
We see that the error dynamics follow a similar trend to the case of stationary targets.
Moreover, for all the methods the localization performance degrades as the targets start to move.

\subsection{Generalization Performance}
The experiments reported so far evaluated the networks with the same number of targets they have seen during training.
In this section we investigate the generalization performance of our method when there are more number of targets than what the network was trained on.

Since the image representation is agnostic to the number of targets to be localized, testing our networks on generalized number of targets is seamless.
In this set of experiments, we take our policy network trained with 2 targets and evaluate it on $m = 6$ and $12$ targets. The results presented in Table~\ref{tab:generalization} suggest that our method generalizes to localizing unseen number of targets, owing to the image representation it uses for the observations.

\begin{table}[ht!]
\centering
\begin{tabular}{l c c} \toprule
\textbf{Method} & $m=6$  & $m=12$\\  \midrule 
Offline-Fisher & \emph{0.315 / 0.067} & \emph{0.286 / 0.038} \\
Greedy-Local & 2.316 / 0.146 & 1.921 / 0.299  \\
Ours (Image) & \textbf{0.729 / 0.202} & \textbf{0.728 / 0.267} \\
\bottomrule
\end{tabular}
\vspace{0.1cm}
\caption{Generalization performance: Evaluation on $m = 6, 12$ targets using a policy network trained to localize 2 targets with bearing measurements}
\label{tab:generalization}
\end{table}

\section{Conclusion}
In this paper we presented a method for localizing multiple targets using bearing or range measurements.
Leveraging recent advances in reinforcement learning and convolutional networks, our method is able to generate trajectories for a robot to localize targets when obtaining noisy relative observations.
We show that our method outperforms a standard greedy approach and performs similarly compared to an offline algorithm which has access to the true position of the targets, without knowing the true target locations.

Future work includes tracking multiple targets that have adversarial movements.
Another interesting research direction is to study the target localization problem in higher dimensions and cluttered environments such as animal tracking with an aerial vehicle in a forest.


\bibliography{references}
\bibliographystyle{unsrt}

\end{document}